\documentclass{article}

\usepackage[final]{neurips_2022}

\usepackage[utf8]{inputenc} 
\usepackage[T1]{fontenc}    
\usepackage{hyperref}       
\usepackage{url}            
\usepackage{booktabs}       
\usepackage{amsfonts}       
\usepackage{nicefrac}       
\usepackage{microtype}      
\usepackage{xcolor}         

\usepackage{graphicx}
\usepackage{subfigure}
\usepackage{algorithm}
\usepackage{algorithmic}
\usepackage{amsfonts}
\usepackage{multirow}
\usepackage{amssymb}
\usepackage{newtxmath}

\def\supat{A$^2$}

\title{\supat{}: Efficient Automated Attacker for Boosting Adversarial Training}

\author{%
Zhuoer Xu$^{1,2}$, 
Guanghui  Zhu$^{1}$\thanks{Corresponding author.} \ ,
Changhua Meng$^{2}$, 
Shiwen Cui$^{2}$, 
Zhenzhe Ying$^{2}$, \\ 
\textbf{Weiqiang Wang$^{2}$, 
Ming Gu$^{2}$, 
Yihua Huang$^{1}$}
\\
\text{$^{1}$State Key Laboratory for Novel Software Technology, Nanjing University \  $^{2}$Tiansuan Lab, Ant Group}\\
\texttt{zhuoer.xu@smail.nju.edu.cn}, \texttt{zgh@nju.edu.cn}, \\
\texttt{\{changhua.mch,donn.csw,zhenzhe.yzz,weiqiang.wwq,guming.mg\}@antgroup.com},\\
\texttt{yhuang@nju.edu.cn}
}

\begin{document}

\maketitle

\begin{abstract}
Based on the significant improvement of model robustness by AT (Adversarial Training), various variants have been proposed to further boost the performance.
Well-recognized methods have focused on different components of AT (e.g., designing loss functions and leveraging additional unlabeled data).
It is generally accepted that stronger perturbations yield more robust models.
However, how to generate stronger perturbations efficiently is still missed.
In this paper, we propose an efficient automated attacker called \supat{} to boost AT by generating the optimal perturbations on-the-fly during training. 
\supat{} is a parameterized automated attacker to search in the attacker space for the best attacker against the defense model and examples.
Extensive experiments across different datasets demonstrate that \supat{} generates stronger perturbations with low extra cost and reliably improves the robustness of various AT methods against different attacks.
\end{abstract}

\section{Introduction}

DNNs (Deep Neural Networks) are extremely vulnerable to imperceptible perturbations despite their success in a wide variety of applications \citep{he2016resnet,kenton2019bert,guo2017deepfm}.
In particular, adding small but carefully chosen deviations to the input, called adversarial perturbations, can cause DNNs to make incorrect predictions with high confidence.
It indicates that models trained by minimizing the empirical risk are not intrinsically robust.
To explicitly improve robustness, AT (Adversarial Training), where a defense model is trained on worst-case adversarial perturbations generated by an attacker, was developed and proved to be effective.

Based on the significant improvement of models' robustness, various methods have been proposed, which focus on different components of AT: analyzing the robustness of neural architectures \cite{Zagoruyko2016wrn}, designing loss functions such as TRADES \citep{zhang2019trades} and MART \citep{wang2019mart}, perturbing the model to regularize the loss landscape's flatness (i.e., AWP \citet{wu2020awp}) and leveraging unlabeled data (i.e., RST \citep{carmon2019rst}).
\citet{gowal2020deepmind} compares the performance of each combination of most components and achieves the SOTA (state-of-the-art) performance.




All the above AT methods use PGD$^{K}$ \citep{madry2018at}, which is a $K$-step stack of FGSM~\citep{fgsm}, as the attacker to generate perturbations for each example against the defense model.
As the key of AT, stronger perturbations yield more robust models.
However, there is a trade-off between the strength of the perturbation and the training efficiency.
Increasing the attack step $K$ strengthens the perturbations, but linearly increases the training overhead~\citep{gowal2020deepmind}.
Likewise, the huge overhead prevents the use of SOTA adversarial attackers \citep{croce2020aa,yao2021aaa}.
To achieve a balance of robustness and efficiency, manually tuning the attacker (e.g., step size and attack method in each step) is of great concern.
R+FGSM \citep{wong2019fastbetterfree} first randomly initializes a small perturbation, and then applies FGSM with the tuned step size.
Surprisingly, AT with R+FGSM is as effective as PGD$^{K}$ but has a significantly lower cost.

However, given the novel dataset, tuning the attacker manually is a challenging task requiring expert knowledge.
Moreover, the best attacker is fine-grained to each example and current model during adversarial training.
Manual coarse-grained tuning for the whole training (e.g., fixed attack method and step size for all examples) is sub-optimal and prevents further improvement of robustness.

Inspired by AutoML (Automated Machine Learning \citep{ZophL17nas,liu2018darts}), we propose an efficient automated attacker called \supat{} to boost AT. 
\supat{} is a parameterized attacker, which can automatically tune itself on-the-fly during training to generate worst-case perturbations for each example.
First, we design a general attacker space by referring to existing attackers.
The attacker space is stacked by one-step attacker cells.
Each cell consists of a perturbation block and a step size block.
Then, we employ a parameterized attacker to search for operations in each block and construct the attacker for each example that maximizes the model loss.
Specifically, we leverage the attention mechanism to calculate the score of each operation.
For continuous operations, we sum up the operations using the normalized scores as weights.
For discrete operations, we use the reparameterization trick~\citep{gumbelsoftmax} to sample an operation from the corresponding block.
In this way, the constructed attacker generates worst-case adversarial perturbations to train the defense model.
Meanwhile, \supat{} is differentiable and can be optimized with respect to the model loss by gradient descent.

We conduct extensive experiments to verify the effectiveness and efficiency of \supat{}.
Compared with PGD, \supat{} can find better perturbations for different models trained with various AT methods.
The results demonstrate that 20-step \supat{} generates better perturbations than PGD$^{100}$.
Moreover, we combine \supat{} with other AT variants, and the robustness of models with different architectures on various datasets is generally improved under strong attacks (e.g., classical C\&W and SOTA AutoAttack).
We also show that \supat{} is insensitive to its hyperparameters.

To summarize, our main contributions can be highlighted as follows:
\begin{itemize}
    \item We propose an efficient automated attacker called \supat{}, which can generate worst-case perturbations on-the-fly during training to improve robustness.
    \item In \supat{}, we design an attacker space by summarizing the existing attackers and employ a differentiable method to construct the most adversarial attacker for each example according to the attention mechanism.
    \item Extensive experimental results across different datasets and neural architectures demonstrate that \supat{} improves the model's robustness by generating stronger perturbations in the \textit{inner maximization}. Moreover, \supat{} can be flexibly combined with different AT methods, showing good generality.
\end{itemize}

\section{Preliminary: Adversarial Training}


Let $D = (\mathbf{X}, Y) = \lbrace (\mathbf{x}_i, y_i) \rbrace_{i = 1}^n$ be a dataset with $\mathbf{x}_i \in \mathbb{R}^d$ as a natural example and $y_i \in \lbrace 1, \ldots, C \rbrace$ as its associated label.
We measure the performance of a DNN classifier $f$ parametrized with $\theta$ using a suitable loss function $l$, denoted as $\mathbb{E}_{(\mathbf{x}_i, y_i) \in D} \left[ l(f_{\theta}(\mathbf{x}_i), y_i) \right]$.
AT \citep{madry2018at} formulates a saddle point problem whose goal is to find the model parameters $\theta$ that minimize the adversarial risk in the \textit{outer minimization} (the example's index $i$ is omitted for brevity):
\begin{equation}
\displaystyle
    \underbrace{
      \mathop{\min}_{\theta} \mathbb{E}_{(\mathbf{x}, y) \in D}
      \overbrace{
        \left[
            \mathop{\max}_{\delta \in \mathbb{S}} l\left(f_{\theta} \left(\mathbf{x} + \delta \right), y \right)
        \right]
      }^{\textit{inner maximization}}
    }_{\textit{outer minimization}}
    \label{eqn:saddle}
\end{equation}
where $\mathbb{S}$ defines the set of allowed perturbations.
The perturbation is usually constrained by $L_p$ norm with a bound $\epsilon$, i.e. $\mathbb{S}=\lbrace \delta | \lVert \delta \rVert_p \leq \epsilon \rbrace$.

The \textit{inner maximization} aims to find an adversarial perturbation against the example that achieves a high loss for the defense model.
However, it is NP-hard to find the optimum of the \textit{inner maximization}.
Various gradient-based attackers have been proposed to approximate its solution, and we classify them according to the number of steps in gradient ascent.
One-step attackers \citep{fgsm,fgm} generate adversarial perturbations as:
\begin{equation}
    \delta^{*} \approx
    \Pi_{\mathbb{S}} \  \eta \cdot \psi \left(
    \nabla_{\mathbf{x}}
    \right)
\end{equation}
where $\nabla_{\mathbf{x}}$ is short for $\nabla_{\mathbf{x}} l\left(f_{\theta} \left(\mathbf{x} \right), y\right)$, $\eta$ is the step size, $\psi$ is a transformation function (e.g., $sgn$ in FGSM and $Identity$ in FGM) and $\Pi$ is the projection.
Since such linearization attacks tend to be trapped in the non-smooth vicinity of the data point, R+FGSM initializes a small \textit{random} perturbation to escape the vicinity and then applies FGSM.
As a typical multi-step attacker, PGD$^{K}$ \citep{madry2018at} can find better perturbations by $K$ step gradient ascent:
\begin{align}
    \mathbf{x}^{(k)} &= \Pi_{\mathbf{x} + \mathbb{S}} \left(
      \mathbf{x}^{(k - 1)} + 
      \eta \cdot \psi \left( \nabla_{\mathbf{x}^{(k-1)}} \right)
    \right)
\end{align}



\section{Methodology}

\subsection{Motivation}


The key of AT is to generate perturbations in the \textit{inner maximization}.
Strong perturbation helps to improve robustness.
It is generally accepted that the step $K$ used to solve the \textit{inner maximization} correlates with the attacker's ability to generate stronger perturbation.
However, larger $K$ leads to a linear increase in training overhead.
\citet{wong2019fastbetterfree} suggests that with appropriate step size tuning and early stopping, one-step attackers yield models with the robustness that is comparable to much more expensive multi-step attackers.
It indicates that hyperparameters, such as random initialization, step size, momentum, and early stopping, affect perturbation generation.
From the perspective of effectiveness and efficiency, it is valuable to further improve robustness by tuning the attacker to strengthen perturbations.

However, manual tuning of perturbation generation for each example on-the-fly during training is impractical.
To address this problem, we propose an \textit{efficient automated attacker} to boost adversarial training by generating optimal perturbations on-the-fly during training.

\subsection{Problem Formulation}
%
Inspired by AutoML, we first design a general attacker space $\mathcal{A}$ by referring to existing attackers.
Then, we employ an \textit{automated attacker} parameterized by $\alpha$ to search in $\mathcal{A}$ and further construct an attacker against the example and the defense model $(\mathbf{x}, y, f_{\theta})$.
%
%
We abbreviate the perturbation generated by the constructed attacker as $\delta_{\alpha}$.
Therefore, the goal of \supat{} is to train a robust model using the perturbation generated by the constructed attackers through a bilevel optimization problem:
\begin{equation}
\label{eqn:bilevel}
    \displaystyle
\begin{split}
&\mathop{\min}_{\theta} \mathbb{E}_{(\mathbf{x}, y) \in D}
    \left[
     l(f_{\theta}(\mathbf{x} + \delta_{\alpha^{*}}), y)
    \right] \\
    \mathrm{s.t.} \  & \alpha^{*} = \mathop{\arg \max}_{\alpha} \mathbb{E}_{\left(\mathbf{x}, y \right) \in D}
    \left[
      l\left(f_{\theta} \left(\mathbf{x} + \delta_{\alpha} \right), y \right)
    \right]
\end{split}
\end{equation}
On the attack side, we train $\alpha$ by SGD to make the defense model misclassify.
On the defense side, we use $\alpha^{*}$ to construct the best attacker for each example and then generate perturbations to adversarially train the defense model.

\begin{figure*}[h]
    \centering
    \subfigure[]{
        \includegraphics[width=0.58\textwidth]{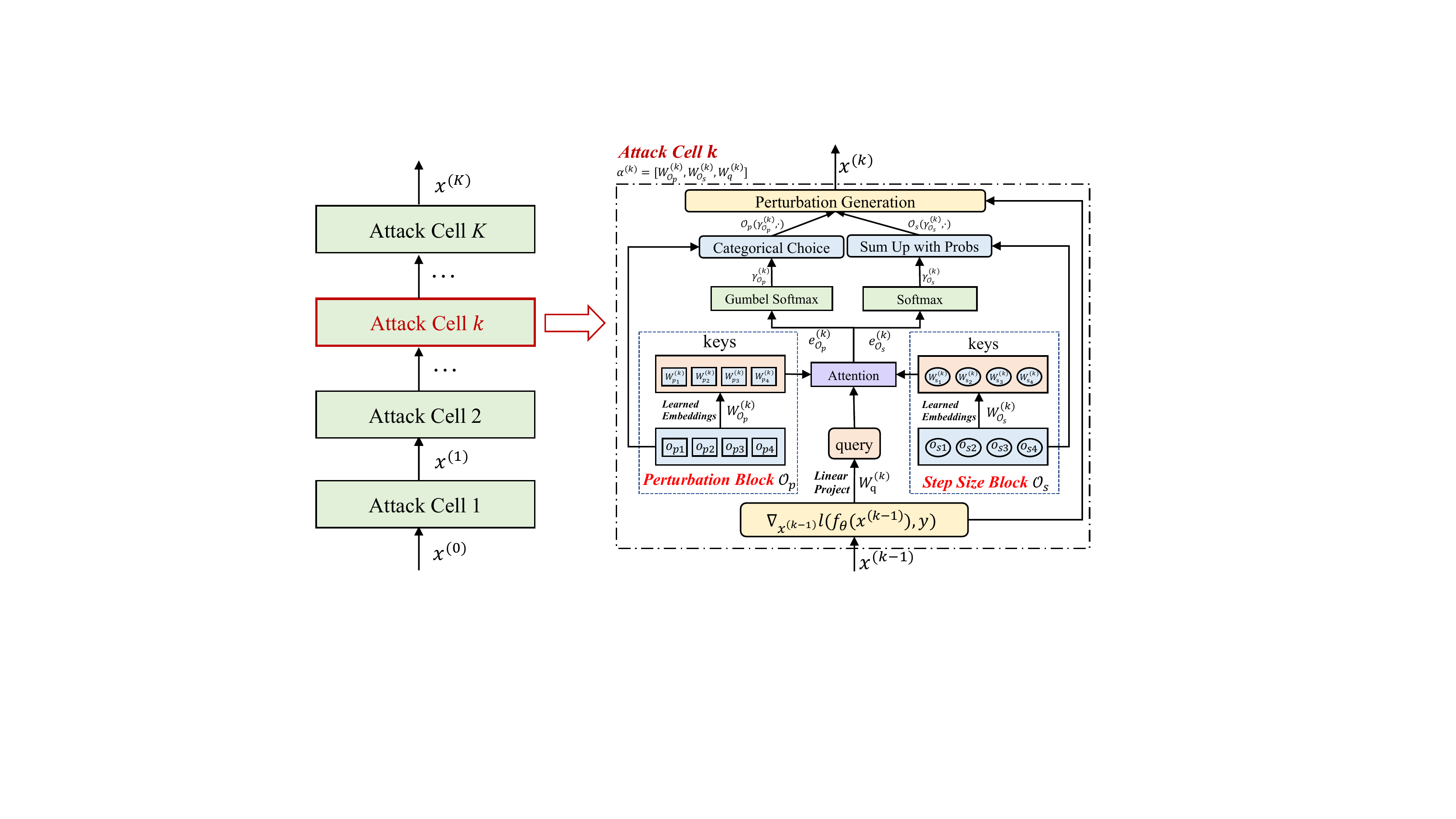}
        \label{fig:search-space}
    }
    \subfigure[
    ]{
        \includegraphics[width=0.2\textwidth]{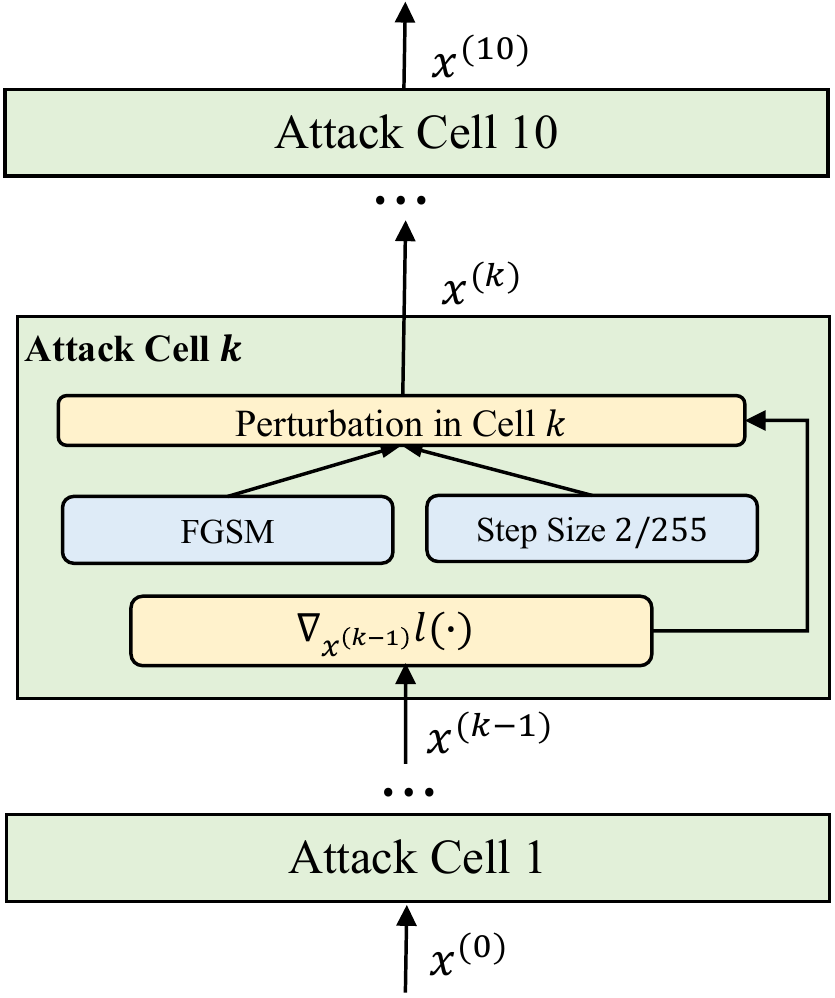}
        \label{fig:pgd10}
    }
    \subfigure[]{
        \includegraphics[width=0.17\textwidth]{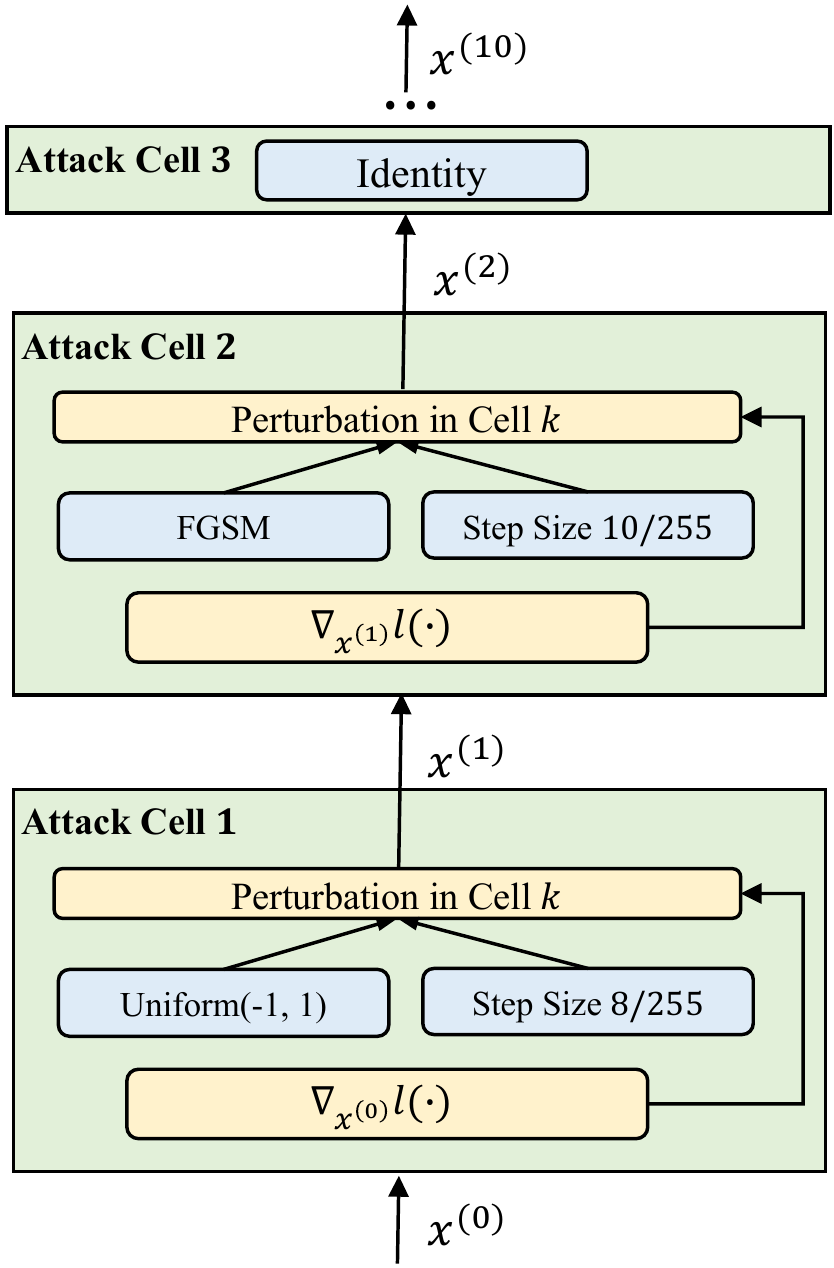}
        \label{fig:r+fgsm}
    }
\caption
{
(a) Attacker Space of \supat{};
(b) PGD$^{10}$: \textit{FGSM} and a fixed step size $2/255$ in each cell;
(c) R+FGSM: \textit{Gaussian} in the first cell, \textit{FGSM} in the second cell and \textit{Identity} in the other cells.
}
\label{fig:overview}
\end{figure*}

\subsection{Attacker Space}


Revisiting most attackers, we find that the attacker can be viewed as a stack of one-step attackers consisting of an attack method and a step size.
Thus, as shown in Figure~\ref{fig:search-space}, we design a general attacker space $\mathcal{A}$ consisting of $K$-step cells.
The $k$-th cell is denoted as $\mathcal{C}^{(k)}$, which is a one-step attacker consisting of the following two blocks:

\textbf{Perturbation Block $\mathcal{O}_{p}$.} Typical attack methods (i.e., \textit{FGM} and \textit{FGSM}), attack methods with momentum (i.e., \textit{FGMM} and \textit{FGSMM}), random perturbations (i.e., \textit{Gaussian} and \textit{Uniform}),
and the special \textit{Identity} which enables the attacker to automatically early stop at a certain step like FAT \citep{zhang2020fat};

\textbf{Step Size Block $\mathcal{O}_{\textit{s}}$.} $\lbrace 10^{-4} \cdot \eta, 10^{-3} \cdot \eta, 10^{-2} \cdot \eta, 10^{-1} \cdot \eta, \eta \rbrace$,
where $\eta$ is a hyperparameter related to the space of allowed perturbations $\mathbb{S}$.

Each block $\mathcal{O}$ contains multiple operations.
Let  $o(\cdot)$ denote the operation, $\gamma^{(k)}=[\gamma_{\mathcal{O}_p}^{(k)}, \gamma_{\mathcal{O}_s}^{(k)}]$ denote the choice of operation in the $k$-th cell.
The attack methods within $\mathcal{O}_p^{(k)}$ are mutually exclusive.
Thus, $\gamma_{\mathcal{O}_p}^{(k)}$ is a one-hot vector. 
In contrast, the operations within $\mathcal{O}_s^{(k)}$ are continuous.
$\gamma_{\mathcal{O}_s}^{(k)}$ is a normalized continuous vector, where each element represents the selection probability.
To unify the categorical choice of attack methods and the probabilities over step sizes, the output of $\mathcal{O}^{(k)}$ is expressed as a mixture based on $\gamma_{\mathcal{O}}^{(k)}$:
\begin{equation}
    \bar{\mathcal{O}} (\gamma_\mathcal{O}^{(k)}, \nabla_{\mathbf{x}^{(k-1)}}) = \sum_{o \in \mathcal{O}^{(k)}}
      \gamma_{o} \cdot o(\nabla_{\mathbf{x}^{(k-1)}})
\end{equation}
where $\gamma_{o}$ denotes the weight of the operation $o$ in $\gamma_{\mathcal{O}}^{(k)}$.
Correspondingly, the one-step attacker of the $k$-th cell can be expressed as the joint of two blocks:
\begin{small}
\begin{align}
    & \bar{\mathcal{C}} (\gamma^{(k)}, \nabla_{\mathbf{x}^{(k-1)}}) =
    \bar{\mathcal{O}}_s (\gamma^{(k)}_{\mathcal{O}_s}, \nabla_{\mathbf{x}^{(k-1)}})
    \cdot
    \bar{\mathcal{O}}_{p} (\gamma^{(k)}_{\mathcal{O}_p}, \nabla_{\mathbf{x}^{(k-1)}})
\end{align}
\end{small}
Moreover, the constructed attacker is a composition of attackers from each cell.
In this way, we can cover common attackers in our space.
For example, as shown in Figure~\ref{fig:pgd10}, PGD$^{K}$ is obtained by selecting \textit{FGSM} in each perturbation block.
R+FGSM in Figure \ref{fig:r+fgsm} is a case of selecting \textit{Gaussian} in the first cell, \textit{FGSM} in the second cell and \textit{Identity} in the other cells.

\textbf{Analysis of $\mathcal{A}$}
Considering there exist $7$ attack methods in $\mathcal{O}_p$ of each step, there are $7^K$ combinations of attack methods in the $K$-step attacker space.
The exponential increasing combinations prevent the brute-force search.
Moreover, the continuous step size $\mathcal{O}_s$ is also part of the attacker space.
Thus, we propose \supat{} to search for the best attacker in $\mathcal{A}$ and generate adversarial perturbations efficiently.

\subsection{Automated Attacker \supat{}}

\supat{} is used to construct the best attacker against $(\mathbf{x}, y, f_{\theta})$ and its trainable parameters $\alpha$ include $W_{\mathcal{O}_p}^{(k)}$, $W_{\mathcal{O}_s}^{(k)}$, and $W_{q}^{(k)}$ where $k\in\{1, \ldots, K\}$.
As shown in Figure~\ref{fig:search-space}, we treat the current model and example as a query and the candidate operations as keys.
Thus, the attention mechanism can be used to calculate the scores of operations within each block and the operations are selected based on their scores.
Specifically, in the $k$-th cell, we take the gradient of the last step $\nabla_{\mathbf{x}^{(k-1)}} l\left(f_{\theta} \left(\mathbf{x}^{(k-1)} \right), y\right)$ as input and project it to a vector space as the query using $W_{q}^{(k)}$.
%
%
%
Then, we use the trainable embedding table $W_{\mathcal{O}}^{(k)}$ to convert the individual operations within $\mathcal{O}^{(k)}$ to continuous keys.
With the Scaled Dot-Product Attention \citep{vaswani2017attn}, we compute the dot products of the query with each key as the score of the operation $o \in \mathcal{O}$ in the $k$-th cell:
\begin{equation}
    e_o^{(k)} = (\nabla_{\mathbf{x}^{(k-1)}} W_{q}^{(k)})^T W^{(k)}_o
\end{equation}

\textbf{Perturbation Block.}
The operations within $\mathcal{O}_{\textit{p}}^{(k)}$ are mutually exclusive.
We sample an operation with the normalized scores as probabilities, i.e., $\gamma_{\mathcal{O}_p}^{(k)} \sim \textit{softmax}(e_{\mathcal{O}_p}^{(k)})$.

\textbf{Step Size Block.}
As the operations within $\mathcal{O}^{(k)}_{\textit{s}}$ are continuous values, we sum up the individual step sizes with the normalized scores as weights. For $o_s \in \mathcal{O}_s^{(k)}$, the weight can be expressed as:
\begin{equation}
\gamma^{(k)}_{o_s} =
    \frac{
      \exp{(e^{(k)}_{o_s})}
    }{
        \sum_{o^{'} \in \mathcal{O}^{(k)}_{\textit{s}}}
        \exp{(e^{(k)}_{o^{'}})}
    }
\end{equation}

\subsection{Training of Automated Attacker}

As mentioned above, we train \supat{} to minimize the following objective by gradient descent:
\begin{equation}
\alpha^{*} = \mathop{\arg \min}_{\alpha} -\mathbb{E}_{\left(\mathbf{x}, y \right) \in D}
\left[
  l\left(f_{\theta} \left(\mathbf{x} + \delta_{\alpha} \right), y \right)
\right]
\end{equation}
However, as a result of constructing the attacker by sampling in each perturbation block, the gradient of the loss w.r.t $\gamma_{\mathcal{O}_p}$ is zero.
To train $\alpha$, we use the reparameterization trick \citep{kingma2013vae} to transfer the randomness of sampling to the auxiliary noise and reformulate the objective function.
For brevity, we omit the step index $k$.

Let $\gamma_{\mathcal{O}_p} = \phi(\kappa, e_{\mathcal{O}_p})$ be a differentiable transformation where $\kappa$ is an auxiliary noise variable with independent marginal $p(\kappa)$.
In \supat{}, we sample noise from Gumbel Distribution, i.e., $\kappa \sim \textit{Gumbel}(0)$ \citep{gumbel1954statistical}, and use Gumbel Softmax \citep{gumbelsoftmax} as $\phi$ to smoothly approximate the expectation of loss \citep{maddison2014gumbelmax}.
Specifically, $\phi (\kappa, e_{\mathcal{O}_p}) = \textit{softmax}\left((e_{\mathcal{O}_p} + \kappa) / \tau \right)$ where $\tau$ is the temperature parameter. When $\tau \rightarrow 0$, the generated samples have the same distribution as $\textit{one\_hot}(\mathop{\arg \max}_{o_p \in \mathcal{O}_p} (e_{o_{p}} + \kappa_{o_p}))$.

Using the reparameterization trick, we can now form MC (Monte Carlo) estimates of the expectation of \supat{}'s loss $l$ for each example, which is differentiable, as follows:
\begin{equation}
\displaystyle
\begin{split}
    & l\left(
            f_{\theta} (\mathbf{x} + \delta_{\alpha}), y
        \right) \\
     =& \mathbb{E}_{\gamma_{\mathcal{O}_p} \sim \textit{softmax}(e_{\mathcal{O}_p})} \left[
     l\left(
            f_{\theta} (
            \mathbf{x} + \bar{\mathcal{C}} ([\gamma_{\mathcal{O}_p}, \gamma_{\mathcal{O}_s}], \nabla_{\mathbf{x}}), y
        \right)
    \right]
        \\
    =& \mathbb{E}_{p(\kappa)} \left[
     l\left(
            f_{\theta} (
            \mathbf{x} + \bar{\mathcal{C}} ([\phi(\kappa, e_{\mathcal{O}_p}), \gamma_{\mathcal{O}_s}], \nabla_{\mathbf{x}}), y
        \right)
    \right] \\
    \approx &\frac{1}{M} \sum_{m=1}^{M}
        l \left(
            f_{\theta} \left(
                \mathbf{x} + \bar{\mathcal{C}}
                (
                    [
                        \phi(\kappa^{(m)}, e_{\mathcal{O}_p}),
                        \gamma_{\mathcal{O}_s}
                    ],
                \nabla_{\mathbf{x}}
                )
            \right), y
            \right)
\end{split}
\label{eqn:mc}
\end{equation}
where $M$ is the number of samples.
In practice, $M=1$ can achieve good performance.
In this way, we reformulate the MC approximation in Equation~\eqref{eqn:mc} of $l$ as the objective function $\hat{l}$. 

Moreover, training $\alpha$ to convergence in each epoch can be prohibitive due to the expensive inner maximization in Equation~\eqref{eqn:bilevel}.
We use a simple approximation scheme following the common methods \citep{finn2017maml,liu2018darts}:
\begin{equation}
    \alpha^{*} \approx \alpha + \xi \nabla_{\alpha} \mathbb{E}_{\left(\mathbf{x}, y \right) \in D}
    \left[
      \hat{l}\left(f_{\theta} \left(\mathbf{x} + \delta_{\alpha} \right), y \right)
    \right]
    \label{eqn:one-step}
\end{equation}
where $\alpha$ denotes the current weights of the attacker and $\xi$ is the learning rate.

\begin{algorithm}[h]
\caption{Adversarial Training with Automated Attacker (AT-\supat{})}
\label{alg:a2t}
\textbf{Input}: Training examples $D$, perturbation bound $\epsilon$, the number of attack steps $K$

\begin{algorithmic}[1] 

\STATE Initialize $\theta, \alpha$;

\FOR{epoch = $1, \ldots ,N_{ep}$}

    \FOR{minibatch $(\mathbf{X}, Y) \subset D$}
        \STATE $\mathbf{X}^{(0)} \leftarrow \mathbf{X}$;
        \FOR{k = $1, \ldots ,K$} \label{alg:l:start}
        \STATE Calculate the gradient $\nabla_{\mathbf{X}^{(k - 1)}}$;
        \STATE Construct $\delta^{(k)}_{\alpha} \in \mathbb{S}$ according to $\nabla_{\mathbf{X}^{(k - 1)}}$ by $g_{\alpha}$;\label{alg:l:diff}
        \STATE $\mathbf{X}^{(k)} = \mathbf{X}^{(k-1)} + \delta^{(k)}_{\alpha}$;
        \ENDFOR\label{alg:l:end}
        \STATE Update $\theta$ with $\nabla_{\theta} \sum_{(\mathbf{x}, y)} l\left(f_{\theta}(\mathbf{x}^{(K)}), y\right) $;\label{alg:l:loss} 
        \STATE Update $\alpha$ by Equation~\eqref{eqn:one-step};
    \ENDFOR
\ENDFOR

\end{algorithmic}
\end{algorithm}

\subsection{Framework of Adversarial Training with \supat{}}

The overall procedure is shown in Algorithm \ref{alg:a2t}.
As in normal adversarial training, we generate perturbations in $K$ steps every batch and update the model parameters.
The key difference is in Line \ref{alg:l:diff}.
Benefiting from a parameterized automated attacker, we tune the discrete attack methods and continuous step sizes to generate adversarial perturbations.
After optimizing the model parameters, we use Equation~\eqref{eqn:one-step} to update $\alpha$ as an approximation to $\alpha^{*}$.
Since \supat{} focus on the \textit{inner maximization}, it can be compatible with most adversarial training methods.
For example, it is flexible to use the loss function of TRADES or MART for \textit{outer minimization} in Line \ref{alg:l:loss} (i.e., TRADES-\supat{} and MART-\supat{}), or include early stopping in Line \ref{alg:l:start}$\sim$\ref{alg:l:end} as FAT.

\section{Experiments}

We conduct extensive experiments on public datasets to answer the following questions:
1) Can \supat{} generate stronger adversarial perturbations?
2) How effective is the adversarial training with \supat{}?
3) Is \supat{} robust to hyperparameters?
All experiments are run using GeForce RTX 3090 (GPU) and Intel(R) Xeon(R) Silver 4210 (CPU) instances.

\subsection{Effectiveness of Automated Attacker (RQ1)}

In this part, we fix the model $f_{\theta}$, train the automated attacker alone and investigate whether \supat{} can generate more powerful perturbations compared to the commonly used PGD.

\begin{figure}[h]
    \centering
    \subfigure{
        \includegraphics[width=0.7\textwidth]{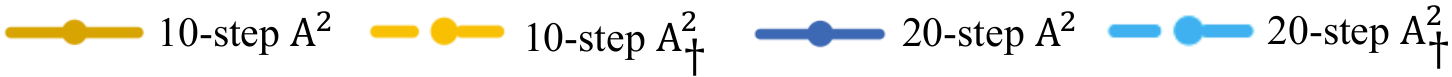}
    }
    \quad
    \setcounter{subfigure}{0}
    \subfigure[ResNet-18 trained by MART]{
        \includegraphics[width=0.48\textwidth]{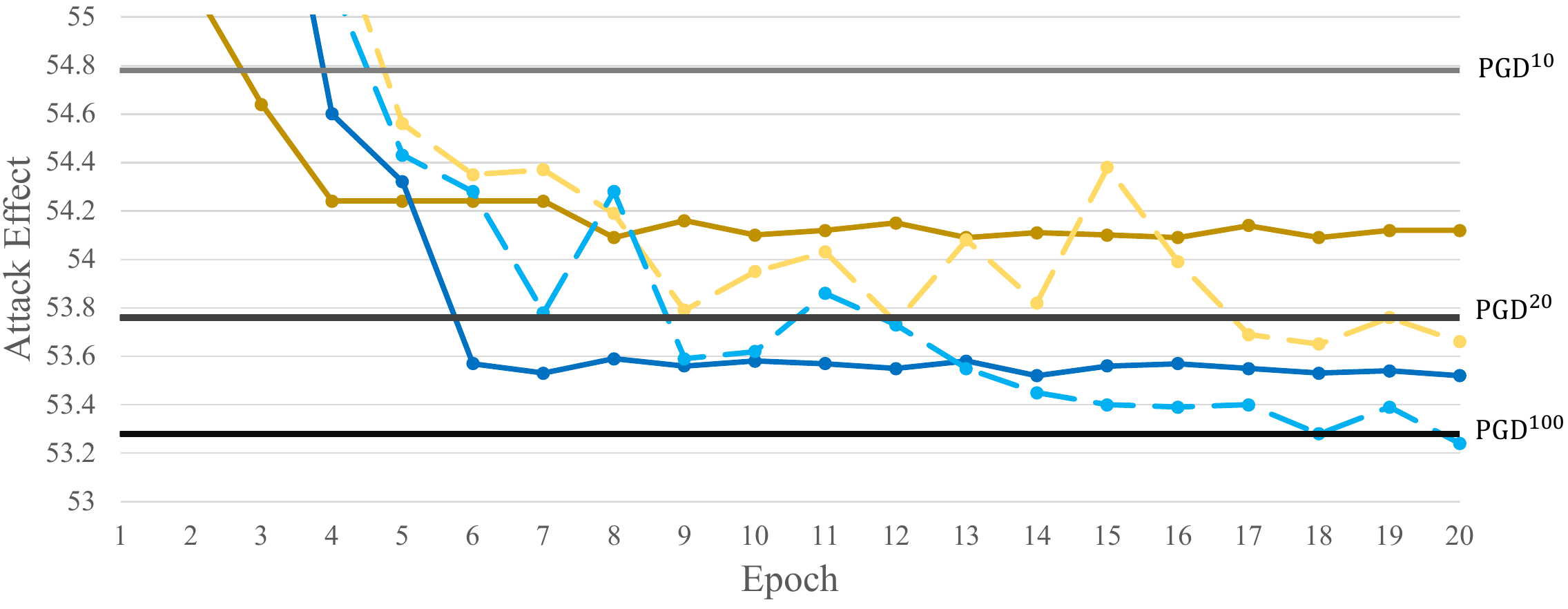}
    }
    \subfigure[WRN-34-10 trained by TRADES-AWP]{
        \includegraphics[width=0.48\textwidth]{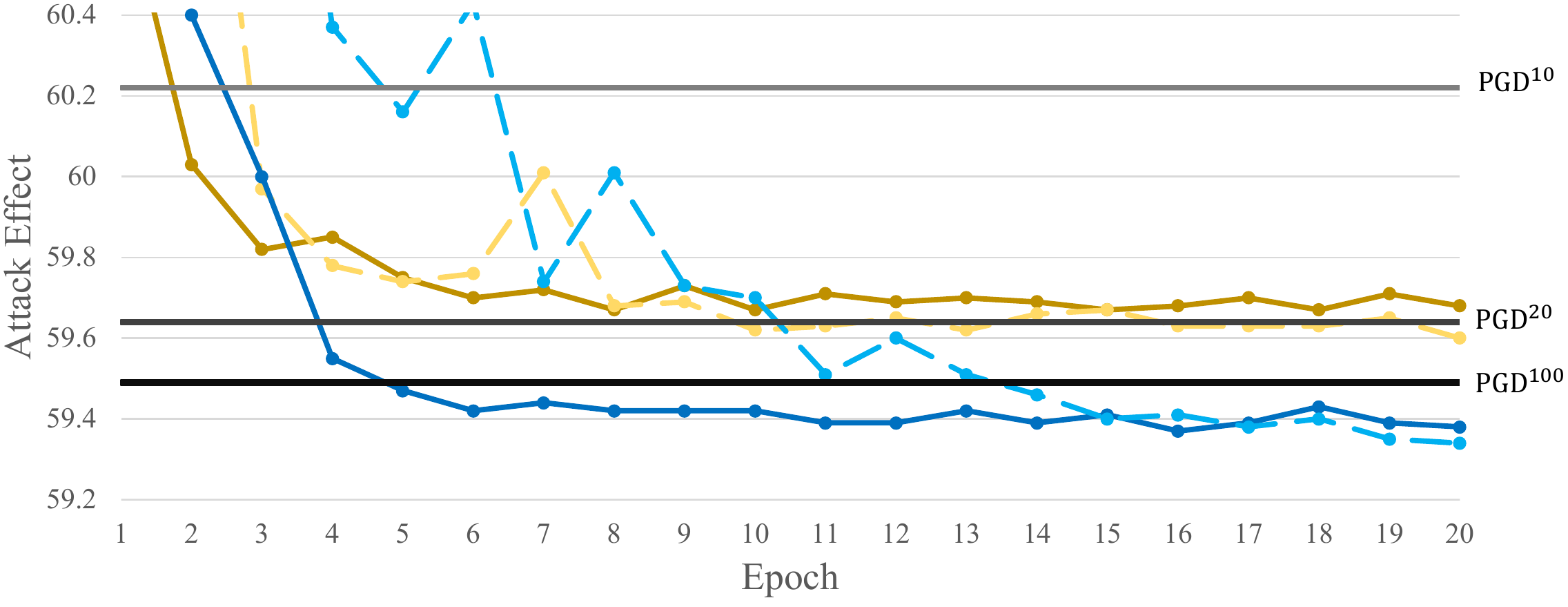}
    }
    \quad
    \subfigure[WRN-34-10 trained by MART-AWP]{
        \includegraphics[width=0.48\textwidth]{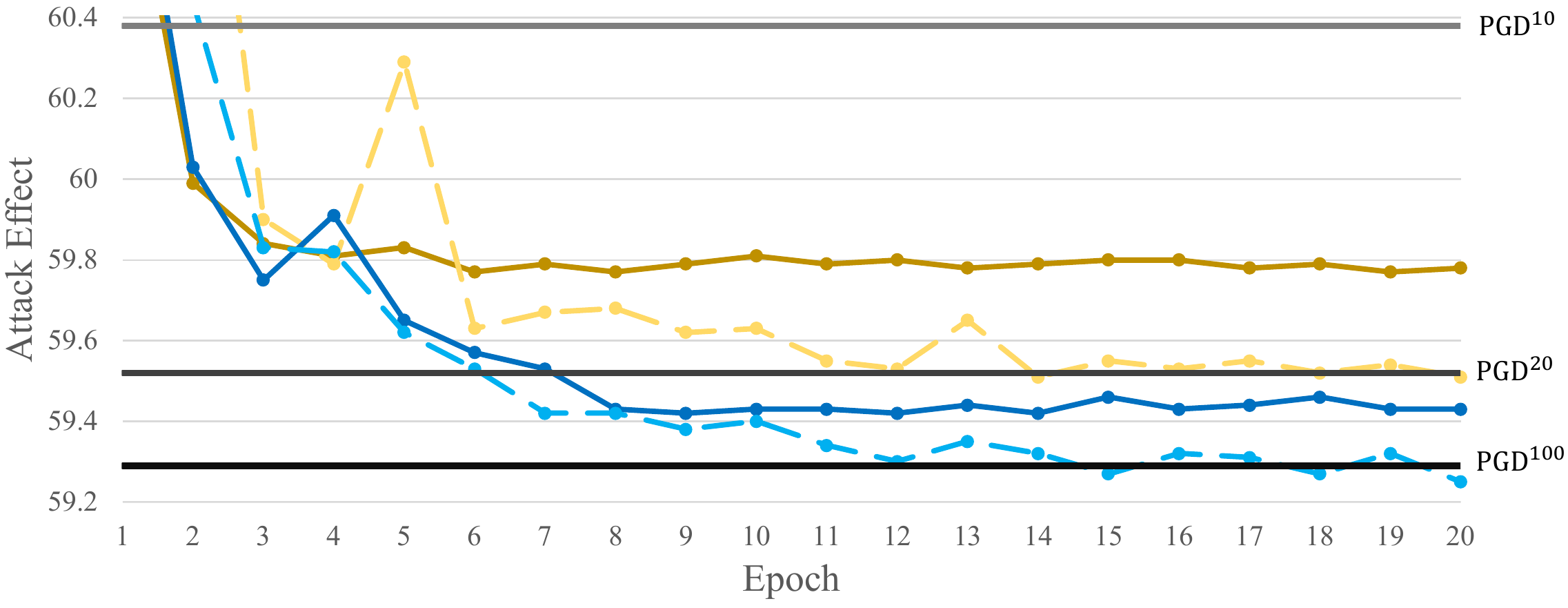}
    }
    \subfigure[WRN-28-10 trained by RST-AWP]{
        \includegraphics[width=0.48\textwidth]{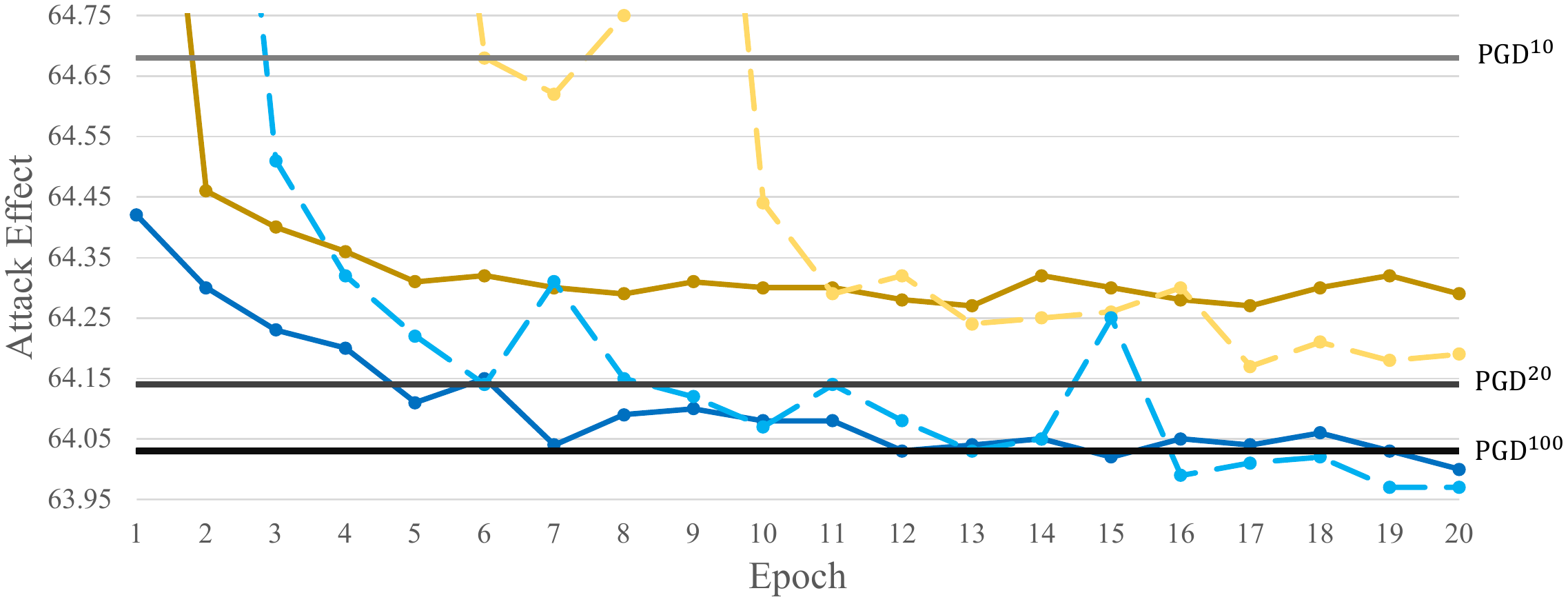}
     }
\caption
{
Effect of adversarial perturbations generated by \supat{} with the training epoch.
}
\label{fig:attack-epoch}
\end{figure}

\begin{table}[htb]
\caption{
Comparison of attack effects ($\%$, the lower the better) of multi-step PGD and \supat{} in robust models.
We run each method 5 times and show the average. The standard deviations are omitted as they are very small.
The architecture of all defense models is WideResNet, except for MART whose architecture is ResNet-18.
}
\centering
\label{tab:attack}
\begin{tabular}{@{}lllllllll@{}}
\toprule
        &         & \multicolumn{3}{c}{10-step} & \multicolumn{3}{c
        }{20-step} &      \\
\cmidrule(r){3-5}     \cmidrule(r){6-8}
Defense     & Natural & PGD & \supat{} & \supat{}$_{\dag}$  & PGD & \supat{} & \supat{}$_{\dag}$ & PGD$^{100}$ \\ \midrule
MART\footnotemark[0]          & {83.07}   & 54.78  & 54.09 &53.65  & 53.76 & 53.52 & \textbf{53.24}  & 53.28  \\
TRADES-AWP\footnotemark[1]   & {85.36}   & 60.22 & 59.67 & 59.60 & 59.64 & 59.38  & \textbf{59.34} & 59.49  \\
MART-AWP\footnotemark[1]     & {85.60}   & 60.38 & 59.76 & 59.51 & 59.52 & 59.42 & \textbf{59.25} & 59.29  \\
RST-AWP\footnotemark[1]      & {88.25}   & 64.68 & 64.27  & 64.17 & 64.14 & 64.02 & \textbf{63.97} & 64.03\\ \bottomrule
\end{tabular}
\end{table}

\textbf{Experimental Settings.}
To demonstrate the generality, we choose different neural architectures (i.e., ResNet-18, WRN-34-10, and WRN-28-10) trained on CIFAR-10 by various AT methods:
TRADES, MART, RST, and AWP.
All trained models are open-source checkpoints.
We choose PGD$^{K}$ with a random start $\delta^{(0)} \sim Uniform(-\epsilon, \epsilon)$ as baseline.
All attacks are $L_{\infty}$-bounded with a total perturbation scale of $\epsilon=8/255$.
Since more attack steps generally improve the attack effect, we compare the automated attacker with PGD at different steps.
Moreover, we try different step size blocks, i.e., $\eta$, in \supat{}: 1) $\eta=2/255$, which is the same as the setting of step size in PGD; 2) $\eta=8/255$, which is indicated by \supat{}$_{\dag}$ and allows \supat{} to search the whole $\epsilon$ bound each step.
\supat{} is trained using Adam \citep{Kingma2015adam} with learning rate $10^{-3}$, weight decay $10^{-2}$ and other default hyperparameters.
\footnotetext[0]{https://github.com/YisenWang/MART}
\footnotetext[1]{https://github.com/csdongxian/AWP}
\footnotetext[2]{https://github.com/zjfheart/Friendly-Adversarial-Training}

\textbf{Attack Effect.}
The training process of \supat{} is shown in Figure~\ref{fig:attack-epoch}.
We take the first 20 epochs of the attack effects, compare them with PGD, and observe whether \supat{} converges.
In the early training stage, the random combinations of attack operations are much less effective.
After 10$\sim$20 epochs, the effect of generated attacks is much more stable and effective.
In practical automated adversarial training, the model and the automated attacker are trained iteratively.
The fast convergence of \supat{} ensures that the generated perturbations are strong enough.
In addition, a larger $\eta$ achieves better attacks.
As the steps increase, the effect diminishes and the training may fluctuate.

Table \ref{tab:attack} reports the attack effects of PGD and \supat{} with different steps.
In the comparison of \supat{} and PGD with different steps $K \in \lbrace 10, 20\rbrace$, \supat{} stably outperforms PGD.
\supat{}$_{\dag}$ constructs stronger attacks in the expanded search space.
With the increase of steps, \supat{} is more effective due to the combination of attack methods and the automated step size tuning.
Due to the diminishing marginal effect, PGD$^{100}$ v.s. PGD$^{20}$ achieves less improvement than PGD$^{20}$ v.s. PGD$^{10}$.
At $1/5$ of the cost, the 20-step \supat{}$_{\dag}$ finds better attacks using the optimized $\alpha$ than PGD$^{100}$.
%
%
%
In summary, Table \ref{tab:attack} verifies that \supat{} stably outperforms PGD for the same step, and obtains better attacks compared to PGD, whose step size and attack method are fixed, with significantly lower cost.

\textbf{Overhead Analysis.} The overhead of \supat{} is not significant compared to PGD. 
Both methods are close in terms of clock
time. 
For WRN-34, PGD takes 19.75/147.09/287.76 seconds
to generate 1/10/20 step attacks respectively. It demonstrates
that more inner steps lead to a linear increase in time.
Meanwhile, \supat{} takes 157.61/302.51 seconds to generate the
10/20 step attack respectively. The main overhead remains
in the forward computation and backward propagation of the
defense model. Moreover, Section A.3 in Appendix shows
that the total parameter size of \supat{} is also acceptable.

\subsection{Effectiveness of Adversarial Training with \supat{} (RQ2)}

In this part, we evaluate the robustness of our proposed AT-\supat{}  on different datasets against white-box and ensemble attacks.
To verify that the stronger attacks generated by \supat{} on-the-fly during training can improve robustness, we consider various adversarial training methods (i.e., AT, TRADES, MART, and AWP) without additional data across different datasets.

\begin{table}[htb]
\centering
\caption{Test robustness ($\%$, the higher the better) using PreActResNet-18 under $L_{\infty}$ threat model ("Best" means the highest robustness while "Last" means the robustness at the last epoch). Std. of 5 runs is omitted due to being small.}
\label{tab:benchmark}
\begin{tabular}{@{}lllllll@{}}
\toprule
\multirow{2}{*}{Defense} & \multicolumn{2}{l}{SVHN} & \multicolumn{2}{l}{CIFAR-10} & \multicolumn{2}{l}{CIFAR-100} \\ \cmidrule(l){2-7} 
                         & Best        & Last       & Best         & Last         & Best          & Last          \\ \midrule
AT                 & 53.36       & 44.49      & 52.79        & 44.44        & 27.22         & \textbf{20.82}         \\
AT-\supat{}                    & \textbf{56.76}       & \textbf{44.75}       & \textbf{52.96}        & \textbf{44.59}         & \textbf{28.14}          & 20.28         \\ \midrule
AWP                   & 59.12       & 55.87      & 55.39        & 54.73        & 30.71         & 30.28         \\
AWP-\supat{}              & \textbf{61.42}       & \textbf{58.45}      & \textbf{55.71}         & \textbf{55.31}         & \textbf{31.36}         & \textbf{30.73}         \\ \bottomrule
\end{tabular}
\end{table}

\textbf{Benchmark.}
\label{sec:benchmark}
We conduct experiments on the baseline AT and the SOTA AWP with \supat{} across three benchmark datasets to verify the generalization of \supat{}.
We follow the settings in AWP: PreActResNet-18 trained for 200 epochs, $\epsilon=8/255$ and $\gamma=10^{-2}$ for AWP.
The step size is $1/255$ for SVHN and $2/255$ for CIFAR-10 and CIFAR-100.
For AT and AWP, the attacker used in training is PGD$^{10}$.
The 10-step \supat{} is trained with the same setting as in RQ1.
PGD$^{20}$ is used for testing, and the test robustness is reported in Table \ref{tab:benchmark}.
It shows that \supat{}, as a component focusing on the \textit{inner maximization}, achieves better results on most datasets.
Moreover, \supat{} is generic and can boost the robustness of both baseline and SOTA AT methods.

\begin{table}
\centering
\caption{Test robustness ($\%$, the higher the better) on CIFAR-10 using WRN-34-10 under $L_{\infty}$ threat model ("Natural" denotes the accuracy on nature examples, and other columns indicate the accuracy on adversarial examples generated by different attacks). Std. of 5 runs is omitted due to being small.}
\label{tab:robust}
\begin{tabular}{@{}llllll@{}}
\toprule
Defense    & Natural & FGSM  & PGD$^{20}$ & CW$_{\infty}$ & AutoAttack \\ \midrule
AT         & \textbf{87.30}   & 56.10 & 52.68 & 50.73  & 47.04 \\
AT-\supat{}    & 84.54   & \textbf{63.72} & \textbf{54.68} & \textbf{51.17}  & \textbf{48.36} \\ \midrule
TRADES    & 84.65   & 61.32 & 56.33 & 54.20 & 53.08 \\
TRADES-\supat{} & \textbf{85.54}   & \textbf{65.93} & \textbf{59.84} & \textbf{56.61} & \textbf{55.03}  \\ \midrule
MART       & 84.17   & 61.61 & 57.88 & 54.58  & 51.10 \\
MART-\supat{}   & \textbf{84.53}  & \textbf{63.73} & \textbf{59.57} & \textbf{54.66}  & \textbf{52.38}\\
\midrule
AWP        & 85.57   & 62.90 & 58.14 & 55.96  & 54.04 \\
AWP-\supat{}        &\textbf{87.54}    & \textbf{64.70}  &\textbf{59.50}  & \textbf{57.42}  & \textbf{54.86}   \\\bottomrule
\end{tabular}
\end{table}

\textbf{Robustness on WideResNet.}
Furthermore, we train WRN-34-10 on CIFAR-10 with various AT methods (i.e., AT, TRADES, MART, and AWP) following their original papers and open-source codes\footnotemark[2].
All defense models are trained using SGD with momentum 0.9, weight decay $5\times 10^{-4}$, and an initial learning rate of 0.1 that is divided by 10 at the $50\%$-th and $75\%$-th epoch.
Except for 200 epochs in AWP, other AT methods train the model for 120 epochs.
Simple data augmentations (i.e., 32x32 random crop with 4-pixel padding and random horizontal flip) are applied.

For white-box attack, we test FGSM, PGD$^{20}$ and CW$_{\infty}$ \citep{carlini2017cw}.
In addition, we test the robustness against the standard AutoAttack~\citep{croce2020aa}, which is a strong and reliable attacker to verify the robustness via an ensemble of diverse parameter-free attacks including three white-box attackers and a black-box attacker.
Table \ref{tab:robust} shows that \supat{} reliably boosts AT variants against white-box and ensemble attacks.
This verifies that \supat{} is general for AT and improves adversarial robustness reliably rather than gradient obfuscation or masking.

Additionally, given the nature examples, AT performs better than AT-\supat{}.
The main reason is that \supat{} generates stronger perturbation for better robustness, which decreases the accuracy (i.e., 84.54). 
Many works (e.g., TRADES, MART, and AWP) use regularization to achieve the trade-off between robustness and accuracy. The regularization is also used to optimize the automated attacker.
Thus, for other AT methods in Table~\ref{tab:robust}, combining \supat{} can achieve higher accuracy.
Moreover, for WRN-34, the training time of AWP-\supat{} is 970 s/epoch while the training time of AWP is 920 s/epoch. Thus, the additional
overhead of \supat{} is not significant.

\subsection{Hyperparameters of \supat{} (RQ3)}

The hyperparameters of \supat{} include the training hyperparameters and the design of the attacker space.
The comparison of the attack effect with different hyperparameters is shown in Table~\ref{tab:hyper}.
Overall, \supat{} is robust to hyperparameters and performs better than PGD$^{10}$ and closely to PGD$^{20}$.


\textbf{Training of \supat{}.}
The effect of attacks with different learning rates $\xi$ is shown in the middle two rows of Table~\ref{tab:hyper}.
Although Adam uses a dynamic learning rate, an excessive initial learning rate (i.e., $10^{-2}$) leads to sub-optimal.

\textbf{Attacker Space.}
The influence of the attack step $K$ has been investigated in RQ1.
As shown in the last two rows of Table~\ref{tab:hyper}, a larger step size $\eta$ increases the effectiveness of \supat{}.
However, as shown in the training curve in Figure~\ref{fig:attack-epoch}, larger $\eta$ introduces instability in the training of the attacker.

\begin{table}[t]
\centering
\caption{Comparison of attack effects ($\%$, the lower the better) of 10-step \supat{} with different hyperparameters (The 10-step is omitted, \supat{}$_{p,q}$ is short for training the attacker using learning rate $\xi=10^{-p}$ with the step size block $\eta=q/255$, ). Std. of 5 runs is omitted due to being small.
}
\label{tab:hyper}
\begin{tabular}{@{}llll@{}}
\toprule
Attack &  TRADES-AWP\footnotemark[1] & MART-AWP\footnotemark[1] & RST-AWP\footnotemark[1] \\ \midrule
PGD$^{10}$          & 60.22    & 60.38  & 64.68   \\
PGD$^{20}$          & 59.64     & 59.52   & 64.14   \\
\supat{}$_{3, 2}$    & 59.67    & 59.76      & 64.27   \\
\midrule
\supat{}$_{2, 2}$    & 59.93    & 59.87      & 64.34   \\
\supat{}$_{4,2}$    & 59.76    & 59.78      & 64.29   \\
\midrule
\supat{}$_{3,5}$    & 59.49    & 59.62      & 64.11   \\
\supat{}$_{3,8}$    & 59.53    & 59.53      & 64.17   \\
\bottomrule
\end{tabular}
\end{table}

\section{Related Work}

\subsection{Adversarial Learning}

Many recent works \citep{fgsm,carlini2017cw,croce2020aa} have shown that DNNs are vulnerable to adversarial examples.
Various defense strategies and models have been proposed to deal with the threat of adversarial examples.
However, as proved in C\&W \citep{carlini2017cw}, many works mistake gradient obfuscation or masking for adversarial robustness.
AT \citep{madry2018at} formulates a class of adversarial training methods for solving a saddle point problem (i.e., Equation~\eqref{eqn:saddle}) and improves robustness reliably.

Based on AT, many works \citep{zhang2019trades,wang2019mart,wu2020awp} focusing on the components of \textit{outer minimization} are introduced to further enhance performance.
The \textit{inner maximization} is also the goal of the adversarial attack, where $l$ is the 0-1 loss.
Many works, e.g., FGSM~\citep{fgsm}, C\&W~\citep{carlini2017cw} and AutoAttack~\citep{croce2020aa}, have been proposed to attack DNNs and facilitate the development of adversarial training.

\subsection{Automated Machine Learning}

AutoML \citep{bergstra2011tpe,ZophL17nas,liu2018darts,cubuk2019autoaugment} aims to automate the parts of the machine learning pipeline that require expert solutions.
For a particular domain, it is common practice to summarize a large search space of parameters and configurations based on expert experience and search for the optimal solutions using methods such as black-box optimization.
During the search process, a certain metric is required to evaluate each solution.
The same idea can be applied to adversarial learning.
$A^3$ \citep{yao2021aaa}, which is also closely related to AutoML, automatically discovers an effective attacker on a given model.

\section{Conclusion}

In this work, we proposed \supat{}, to the best of our knowledge, the first adversarial training method which focuses on automated perturbation generation.
In \supat{}, the attacker space is designed by summarizing the existing perturbations.
Moreover, the parameterized automated attacker leverages the attention mechanism to choose the discrete attack method and the continuous step size and further generates adversarial perturbations.
During training, the one-step approximation of the optimal automated attacker is used to generate the optimal perturbations on-the-fly for the model.
The experimental results show that \supat{} generates stronger attacks with low extra cost and boosts the robustness of various AT methods reliably.

For future work, we plan to add the target loss of the \textit{inner maximization} to the attacker space.
We also plan to apply \supat{} to enhance adversarial training for Natural Language Processing.

\begin{ack}
This work was supported by the National Natural Science Foundation of China (\#62102177 and \#U1811461), the Natural Science Foundation of Jiangsu Province (\#BK20210181), the Key Research and Development Program of Jiangsu Province (\#BE2021729), Ant Group through Ant Research Program, and the Collaborative Innovation Center of Novel Software Technology and Industrialization, Jiangsu, China.
\end{ack}

\bibliographystyle{unsrtnat}
\bibliography{ref}

\section*{Checklist}

\begin{enumerate}

\item For all authors...
\begin{enumerate}
  \item Do the main claims made in the abstract and introduction accurately reflect the paper's contributions and scope?
    \answerYes{}
  \item Did you describe the limitations of your work?
    \answerYes{}
  \item Did you discuss any potential negative societal impacts of your work?
    \answerNo{}
  \item Have you read the ethics review guidelines and ensured that your paper conforms to them?
    \answerYes{}
\end{enumerate}

\item If you are including theoretical results...
\begin{enumerate}
  \item Did you state the full set of assumptions of all theoretical results?
    \answerNA{}
        \item Did you include complete proofs of all theoretical results?
    \answerNA{}
\end{enumerate}

\item If you ran experiments...
\begin{enumerate}
  \item Did you include the code, data, and instructions needed to reproduce the main experimental results (either in the supplemental material or as a URL)?
    \answerYes{}{}
  \item Did you specify all the training details (e.g., data splits, hyperparameters, how they were chosen)?
    \answerYes{}
        \item Did you report error bars (e.g., with respect to the random seed after running experiments multiple times)?
    \answerYes{The standard deviations are omitted as they are very small}
        \item Did you include the total amount of compute and the type of resources used (e.g., type of GPUs, internal cluster, or cloud provider)?
    \answerYes{See Appendix.}
\end{enumerate}

\item If you are using existing assets (e.g., code, data, models) or curating/releasing new assets...
\begin{enumerate}
  \item If your work uses existing assets, did you cite the creators?
    \answerYes{}
  \item Did you mention the license of the assets?
    \answerYes{}
  \item Did you include any new assets either in the supplemental material or as a URL?
    \answerNA{}
  \item Did you discuss whether and how consent was obtained from people whose data you're using/curating?
    \answerNA{}
  \item Did you discuss whether the data you are using/curating contains personally identifiable information or offensive content?
    \answerNA{}
\end{enumerate}

\item If you used crowdsourcing or conducted research with human subjects...
\begin{enumerate}
  \item Did you include the full text of instructions given to participants and screenshots, if applicable?
    \answerNA{}
  \item Did you describe any potential participant risks, with links to Institutional Review Board (IRB) approvals, if applicable?
    \answerNA{}
  \item Did you include the estimated hourly wage paid to participants and the total amount spent on participant compensation?
    \answerNA{}
\end{enumerate}

\end{enumerate}

\appendix

\section{Details in \supat{}}

\subsection{Unify Magnitude of Perturbations}
Perturbations generated by different operations in $\mathcal{O}_{p}$ have different magnitudes and thus require different magnitudes of step size for different $o_p$.
For example, \textit{FGSM} generates perturbations with elements belonging to $\{-1, 0, 1\}$, while the perturbation generated by \textit{FGM} is usually in the magnitude of $10^{-3}$.
Obviously, they cannot use the same step size.
To have a uniform effect of the step size block, we normalize the magnitude of other generated perturbations to be the same as \textit{FGSM} (i.e., $\delta_{o_p} = \delta_{o_p} \cdot \frac{\lVert \delta_{\textit{FGSM}}\rVert}{ \lVert \delta_{o_p} \rVert}$).
In this way, we find that other attack methods such as \textit{FGM} can achieve good results with the same step size as \textit{FGSM}.

\subsection{Temperature Parameter in \textit{Softmax}}

Since there is an order of magnitude difference in step size operations, the larger step size with the same score will dominate the output.
For example, $0.7 \cdot 10^{-2} \eta + 0.3 \cdot \eta \approx 0.3 \eta$.
The output of the step size block is dominated by the operation $\eta$, despite the greater weight of $10^{-2}\cdot \eta$.
To alleviate the problem, we use the temperature parameter $\tau$ in \textit{softmax} to sharpen the distribution:
\begin{equation}
\gamma^{(k)}_{o_{\textit{s}}} =
    \frac{
      \exp{(e^{(k)}_{o_{\textit{s}}} / \tau)}
    }{
        \sum_{o^{'} \in \mathcal{O}_{\textit{s}}}
        \exp{(e_{o^{'}} / \tau)}
    }
\end{equation}
where $o_{\textit{s}}$ is an operation in $\mathcal{O}_{\textit{s}}$, and $e_{o_s}$ is its attention score.
Through experiments, we set $\tau=0.1$ to distinguish the preference for the step size in most cases.

\subsection{Overhead of \supat{}}

Let the number of steps be $K$, the number of operations be $|\mathcal{O}|$, the image size be $W \times H$ and the embedding size be $E$.
The number of the attacker's parameter is $\mathbf{O}\left(K \cdot E \cdot (W\times H + |\mathcal{O}|) \right)$.
Specifically, the number of parameters for the attacker is $7873280$, which is $17\%$ of the model's parameters (i.e., $46160474$).
In each batch, there is only 1 forward calculation of all cells with 1 backpropagation.
In comparison, the model requires $K$ forward calculations with backpropagation.
Therefore, the additional computational overhead from the attacker is not significant in terms of the number of parameters and computations.

Moreover, PGD and \supat{} are close in terms of clock time.
For WRN-34, PGD takes 19.75/147.09/287.76 seconds to generate 1/10/20 step attacks respectively.
It demonstrates that more inner steps lead to a linear increase in time.
Meanwhile, \supat{} takes 157.61/302.51 seconds to generate the 10/20 step attack respectively.
The main overhead remains in the forward computation and backward propagation of the defense model.
For WRN-34, the training time of AWP-\supat{} is 970 s/epoch while the training time of AWP is 920 s/epoch.

In summary, the additional overhead of \supat{} is not significant.

\subsection{Why No Mixture in $\mathcal{O}_p$}

\begin{figure}[h]
    \centering
    \includegraphics[width=0.30\textwidth]{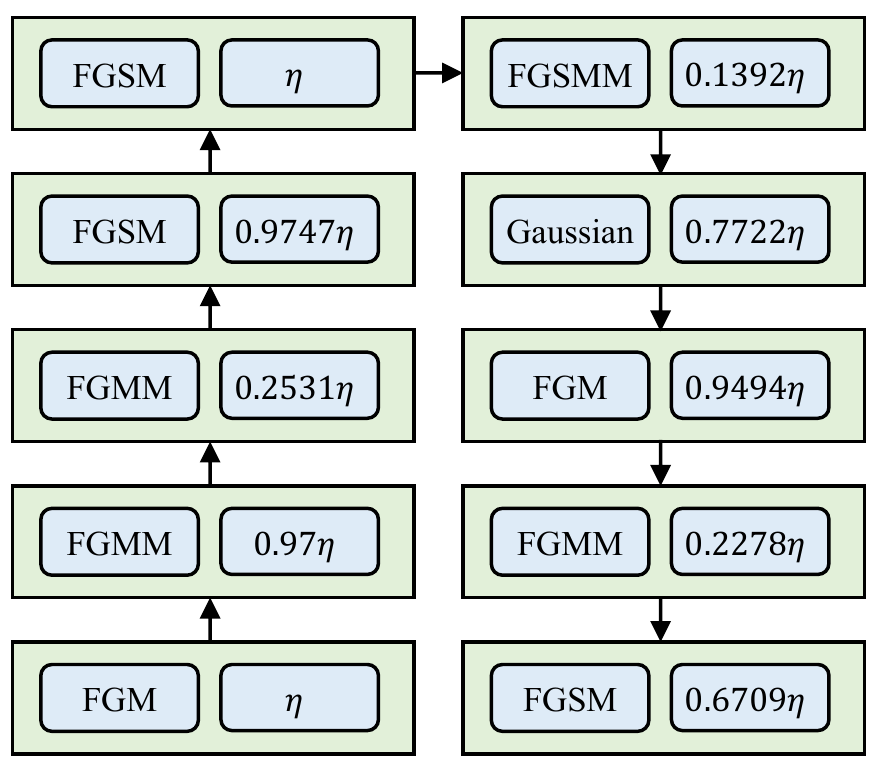}
    \caption{Example of generated attack on CIFAR-10.}
    \label{fig:attack-example}
\end{figure}

Like most NAS methods in AutoML, the discrete selection in the perturbation block is more interpretable and robust (e.g., L1-Norm for feature selection and single path in NAS) than the mixture over possible solutions.
Moreover, the mixture will incur more computational overhead and 7 times memory overhead due to 7 operations in $\mathcal{O}_p$.
Figure \ref{fig:attack-example} shows an example of the generated attack on CIFAR-10, which can be migratable. 

\section{Addition Experiments}

\subsection{Why use FGSM-based PGD in RQ1.}

There are multiple single-step attack methods in $\mathcal{O}_p$ for stacking as PGD, e.g., FGM-based PGD and FGSM-based PGD.
The experimental results of the attack effect of PGD based on these attack methods demonstrate that FGSM-based PGD outperforms the stacking of other operations.
Thus, we choose FGSM-based PGD with a random start $\delta^{(0)} \sim Uniform(-\epsilon, \epsilon)$ as a baseline for comparison with the automated attacker.

\subsection{Number of samples $M$ in MC Approximation.}

$M$ is an important hyperparameter that dictates the quality of MC approximation and the training overhead.
We test the cases with $M \in \{1, 2, 5\}$ and achieve similar performance.
Thus, we set $M$ to $1$ and achieve good results with a significantly lower overhead.

\subsection{Generality of \supat{} in White-Box Attacks}

\begin{table}[htbp]
\caption{Comparison of attack effects on CIFAR-10 ($\%$, the lower the better) of PGD-based and CW$_{\infty}$-based attacks.
The architecture of all defense models is WideResNet, except for MART whose architecture is ResNet-18.}
\centering
\label{tab:cw}
\begin{tabular}{@{}lllll@{}}
\toprule
         & MART  & TRADES-AWP & MART-AWP & RST-AWP \\ \midrule
Natural & 83.07 & 85.36 & 85.60 & 88.25         \\
\midrule
PGD$^{20}$ & 53.76 & 59.64 & 59.52 & 64.14 \\
PGD$^{20}$-\supat{} & 53.24 & 59.34 & 59.25 & 63.97\\
\midrule

CW$_{\infty}$       & 49.97 & 57.07      & 56.44    & 61.82   \\
CW$_{\infty}$-\supat{} & \textbf{49.82} & \textbf{56.98}      & \textbf{55.81}    & \textbf{61.30}   \\ \bottomrule
\end{tabular}
\end{table}

In this part, we investigate whether \supat{} is general to white-box attacks.
As a more powerful attack method, CW$_{\infty}$-based attacks ~\citep{carlini2017cw} stably outperform PGD-based attacks.
For comparison with CW$_{\infty}$, we propose a variant of \supat{} that uses CW$_{\infty}$ loss to generate perturbations and denote it as CW$_{\infty}$-\supat{}.
The results in Table~\ref{tab:cw} show that \supat{} is general and can improve the attack effect of PGD and CW$_{\infty}$ by combining attack methods and tuning the step size.
Moreover, the additional overhead of \supat{} is 5\% to 10\%, which is a rather acceptable trade-off.

\subsection{Robustness Against Transferable Black-Box Attacks}

We investigate the robustness of \supat{} against transferable black-box attacks.
Table~\ref{tab:transfer} provides test robustness on CIFAR-10 using ResNet-18.
%
We adopt three transferable black-box attack methods: MI (momentum = 1)~\citep{dong2018mi}, DI~\citep{xie2019di}, and TI~\citep{dong2019ti}.
The transferable attacks are generated by an ensemble of the above methods on three surrogate pre-trained models~\footnote{https://github.com/huyvnphan/PyTorch\_CIFAR10}: IncV3 (InceptionV3), VGG19, and DN201 (DenseNet201).
Table~\ref{tab:transfer} shows that AT boosts the robustness against transferable black-box attacks, and \supat{} can further improve the adversarial robustness.

\begin{table}[htpb]
\caption{
Test robustness ($\%$, the higher the better) on CIFAR-10 using ResNet-18 against transferable black-box attacks.
}
\centering
\label{tab:transfer}
\begin{tabular}{@{}lllll@{}}
\toprule
 & \multicolumn{3}{c}{MI+DI+TI} &  \\
\cmidrule(r){2-4}
 &  IncV3 & VGG19 & DN201 & PGD$^{20}$  \\ \midrule
ResNet-18 & 16.12 &7.37  & 5.35  & 0.02 \\
ResNet-18-AT & 61.98  & 60.81 & 59.63 &  52.79 \\
ResNet-18-AT-\supat{}&  \textbf{62.79} & \textbf{61.85} & \textbf{60.28} &  \textbf{52.96}\\ \bottomrule
\end{tabular}
\end{table}

\subsection{A Closer Look at Selected Attacks}

\begin{figure}[h]
    \centering
    \subfigure{
        \includegraphics[width=0.7\textwidth]{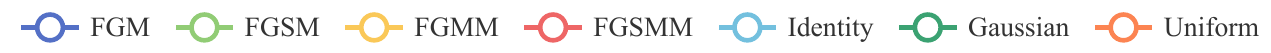}
    }
    \quad
    \setcounter{subfigure}{0}
    \subfigure[Block $1$ in CIFAR-10]{
        \includegraphics[width=0.3\textwidth]{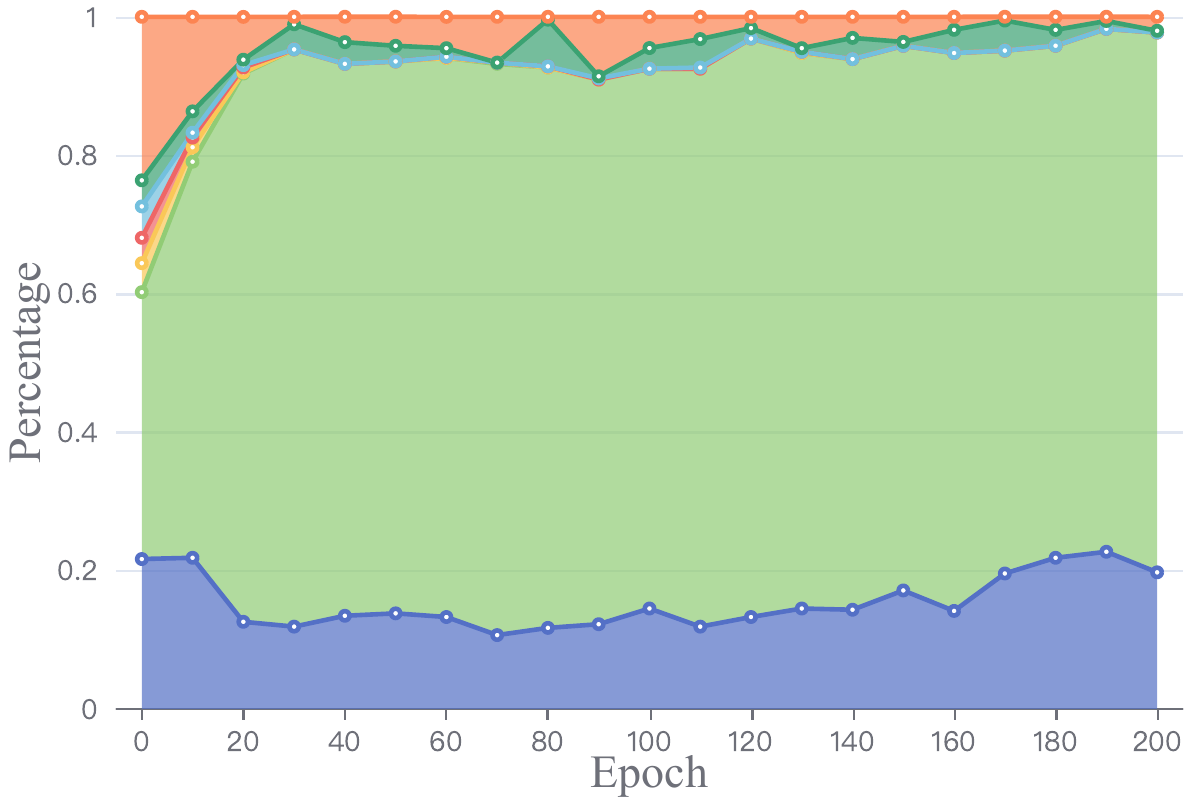}
    }
    \subfigure[Block $10$ in CIFAR-10]{
        \includegraphics[width=0.3\textwidth]{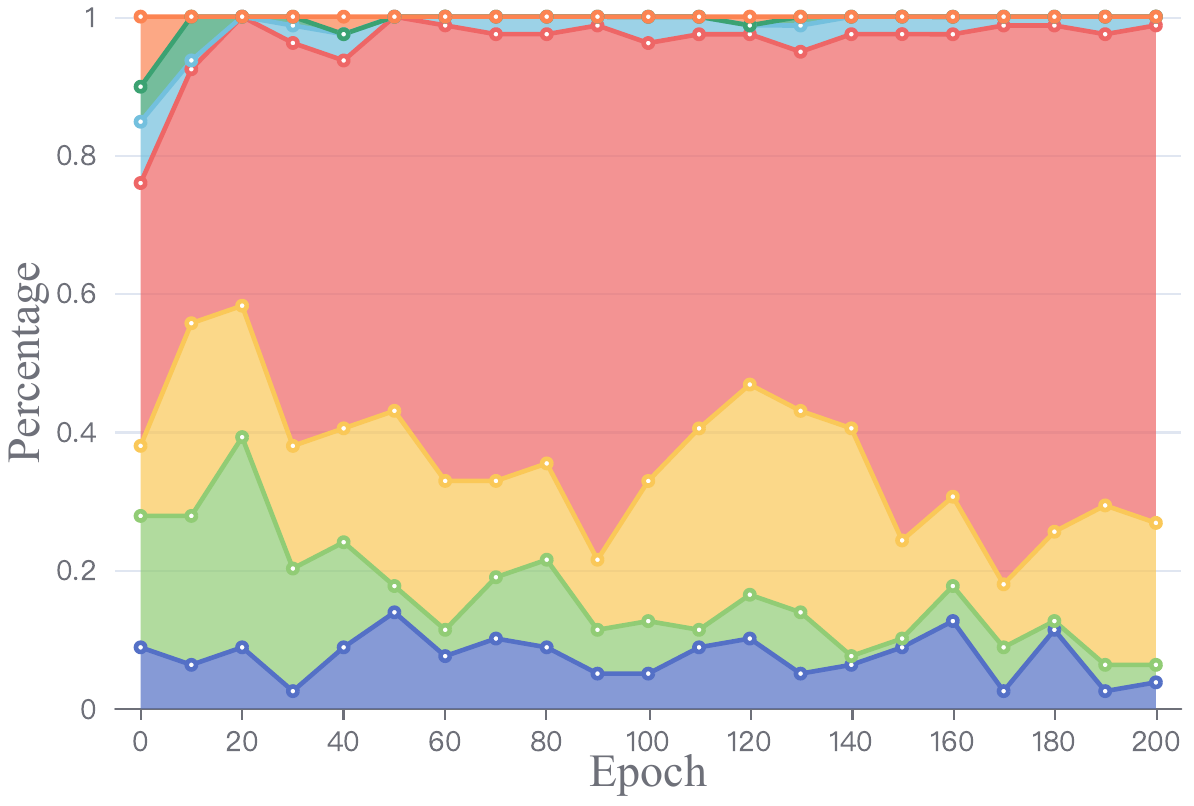}
    }
    \subfigure[Block $10$ in SVHN]{
        \includegraphics[width=0.3\textwidth]{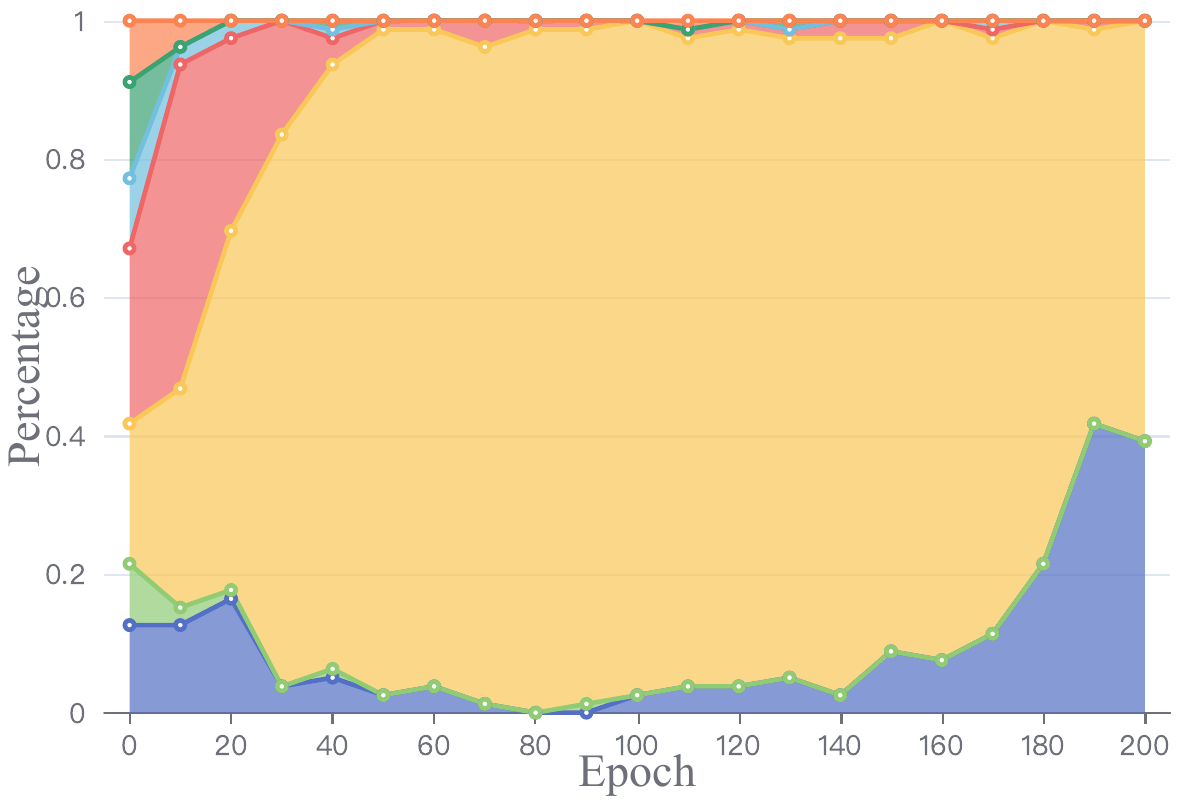}
        \label{fig:svhn}
    }
\caption
{
Distribution of attacks selected by perturbation blocks of \supat{}.
}
\label{fig:step-dist}
\end{figure}

We analyze the selected attacks from the perspective of perturbation blocks with different steps and datasets.

The first and final perturbation blocks of 10-step \supat{} in CIFAR-10 are chosen for analysis.
Figure~\ref{fig:step-dist} shows the distribution of selected attacks of different perturbation blocks.
\begin{itemize}
    \item \textbf{Perturbation Block 1:} \supat{} tends to choose \textit{FGM}, \textit{FGSM}, and partially random methods as initialization in the first step.
    The momentum-based attack methods are quickly discarded as the gradient of the previous step is absent.
    \textit{FGSM} is chosen more frequently due to its stronger attack on both foreground and background.

    \item \textbf{Perturbation Block 10:}
    The optimization of the victim model leads to changes in the distribution of selected attacks in the last block.
    In the early stage of training, the victim model is vulnerable.
    \supat{} retains the diversity and plays the role of friendly attackers like FAT~\citep{zhang2020fat}.
    At the end of training, \supat{} prefers the momentum-based attacks (i.e., \textit{FGSMM} and \textit{FGMM}).
\end{itemize}

From the perspective of datasets, SVHN and CIFAR-10 prefer different attack methods.
As shown in Figure~\ref{fig:svhn}, SVHN discards \textit{FGSMM}, which is most frequently used in CIFAR-10, and pays more attention to \textit{FGMM}.
Moreover, SVHN rarely uses \textit{Identity} compared with CIFAR-10 as its higher robustness accuracy requires more powerful perturbations.


In summary, \supat{}'s preference for selecting attacks in blocks varies according to the block step, dataset, and victim model.

\end{document}